\documentclass[11pt]{article}

\usepackage{acl}

\usepackage{times}
\usepackage{latexsym}
\usepackage{hyperref}
\usepackage{color, colortbl}
\definecolor{Gray}{gray}{0.9}
\definecolor{high}{HTML}{c2d9d6}
\definecolor{low}{HTML}{f9f0c8}
\definecolor{total}{HTML}{dbeda5}
\definecolor{hard}{HTML}{ffebad}
\usepackage{array}
\usepackage{booktabs}
\usepackage{multirow}
\usepackage{amsmath}
\usepackage{amssymb}
\usepackage{graphicx}
\usepackage{enumerate}
\usepackage{diagbox}
\usepackage{mathtools}
\usepackage{paralist}
\usepackage{tabularx}
\usepackage{lipsum}
\usepackage{ragged2e}
\usepackage{makecell}
\usepackage{siunitx}
\usepackage[table]{xcolor}
\usepackage[most]{tcolorbox}

\NewDocumentCommand{\jiayu}
{ mO{} }{\textcolor{blue}{\textsuperscript{\textit{Jiayu}}\textsf{\textbf{\small[#1]}}}}

\usepackage[T1]{fontenc}

\usepackage[utf8]{inputenc}

\usepackage{microtype}

\usepackage{inconsolata}

\usepackage{graphicx}

\title{CritiCal: Can Critique Help LLM Uncertainty or Confidence Calibration?}

\author{
\textbf{Qing Zong, Jiayu Liu, Tianshi Zheng, Chunyang Li, Baixuan Xu, Haochen Shi, } \\ \textbf{Weiqi Wang, Zhaowei Wang, Chunkit Chan, Yangqiu Song} \\
    Department of Computer Science and Engineering, HKUST\\ 
    \texttt{\{qzong, yqsong\}@cse.ust.hk}
  }

\begin{document}
\maketitle
\begin{abstract}

Accurate confidence calibration in Large Language Models (LLMs) is critical for safe use in high-stakes domains, where clear verbalized confidence enhances user trust. 
Traditional methods that mimic reference confidence expressions often fail to capture the reasoning needed for accurate confidence assessment.
We propose natural language critiques as a solution, ideally suited for confidence calibration, as precise gold confidence labels are hard to obtain and often require multiple generations.
This paper studies how natural language critiques can enhance verbalized confidence, addressing: \textit{(1) What to critique}: uncertainty (question-focused) or confidence (answer-specific)? 
Analysis shows confidence suits multiple-choice tasks, while uncertainty excels in open-ended scenarios. 
\textit{(2) How to critique}: self-critique or critique calibration training? 
We propose \textbf{Self-Critique}, enabling LLMs to critique and optimize their confidence beyond mere accuracy, and \textbf{CritiCal}, a novel \textbf{Criti}que \textbf{Cal}ibration training method that leverages natural language critiques to improve confidence calibration, moving beyond direct numerical optimization. 
Experiments show that CritiCal significantly outperforms Self-Critique and other competitive baselines, \textbf{even surpassing its teacher model, GPT-4o}, in complex reasoning tasks.
CritiCal also shows robust generalization in out-of-distribution settings, advancing LLM's reliability.
\footnote{https://github.com/HKUST-KnowComp/CritiCal} 

\end{abstract}

\section{Introduction}

\begin{figure}[t]
     \centering
     \includegraphics[width=0.95\linewidth]{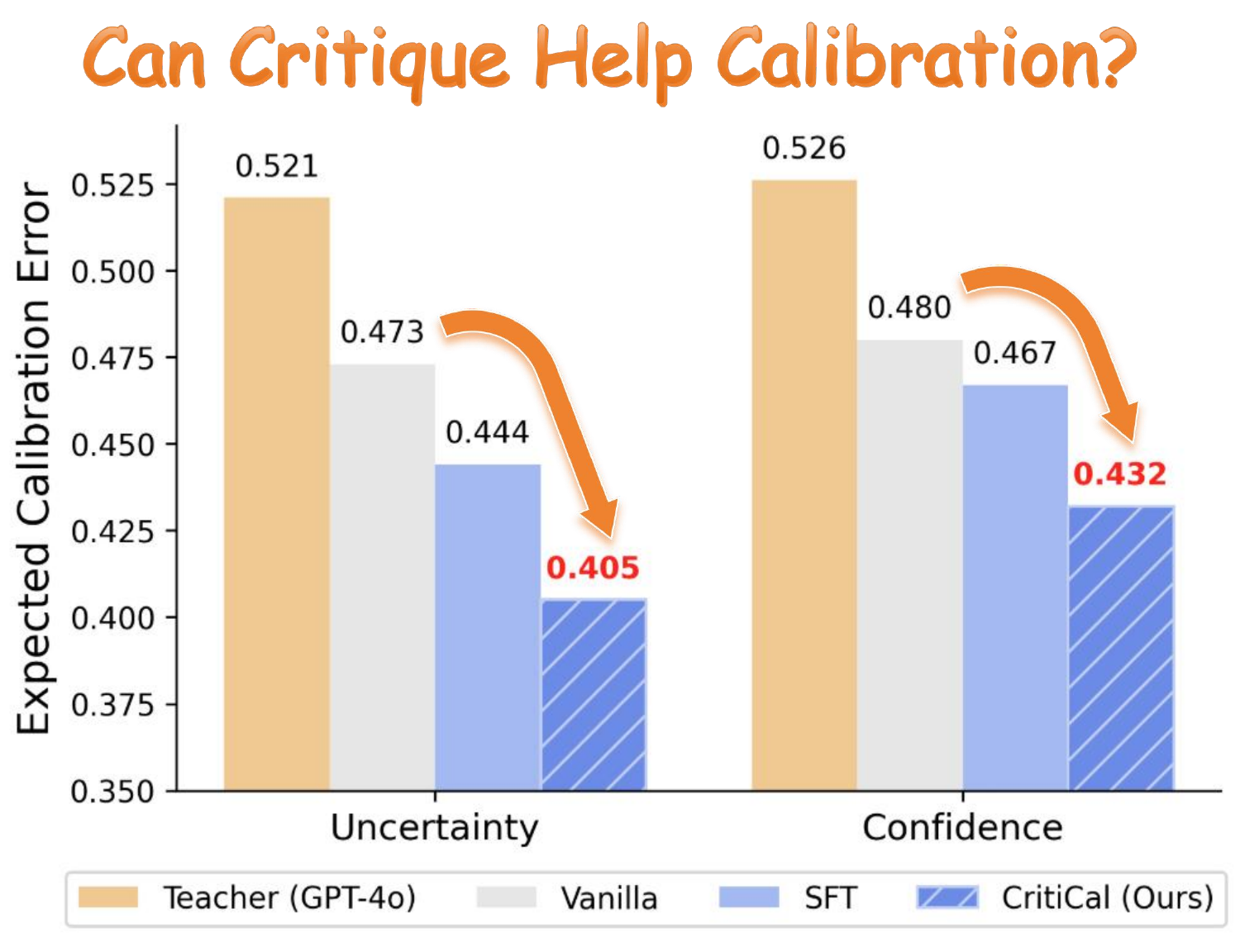}
     \vspace{-0.05in}
     \caption{In-domain comparisons between CritiCal and other SFT methods by DeepSeek-R1-Distill-Qwen-7B on MATH-Perturb, showing CritiCal’s huge potential in improving LLM's confidence calibration even with a teacher model having worse calibration performance.}
    \label{fig:Performance}
    \vspace{-0.2in}
\end{figure}

Confidence calibration is crucial in ensuring the reliability and trustworthiness of LLMs in high-stakes applications~\cite{LM-Polygraph, DBLP:conf/acl/XiaXZL25}. 
As LLMs increasingly interact with humans, verbalized confidence, such as "..., and my confidence is 80\%," allows for clearer communication of response certainty, fostering trust and effective collaboration~\cite{Stephanie, CanLLMsExpress}.

Learning from critique~\cite{cft,critique_grpo} has proven highly effective in improving LLM's accuracy. 
Natural language critiques clarify why answers are correct or incorrect, enabling more reasonable refinements rather than direct imitation of responses. 
This characteristic is ideal for confidence calibration, particularly verbalized confidence, as precise gold confidence labels are hard to obtain, but assessing whether confidence is too high or too low is straightforward based on reasoning and answer correctness. 
However, no related research has been conducted.
This paper bridges this gap by investigating whether critique-based learning can improve uncertainty or confidence calibration, addressing two questions.

\textbf{What to critique: uncertainty or confidence?}
Previous studies often treat uncertainty and confidence as antonyms, overlooking their distinction~\cite{survey}: uncertainty pertains to the whole question, while confidence relates to the specific answer.
Although \citet{DBLP:journals/tmlr/LinT024} explored this difference using a consistency-based method, it was limited to the diversity of model output and did not notice the difference between question types. 
We advance this by conducting a comprehensive study of LLMs' direct outputs and their verbalized uncertainty and confidence. 
For brevity, we use "confidence" to broadly encompass both concepts, distinguishing them only when comparing their specific roles. 
\textit{Extensive experiments show that verbalized confidence excels in multiple-choice questions, while uncertainty is better suited for open-ended tasks.}

\textbf{How to critique: self-critique or critique calibration training?} 
Unlike prior self-improvement methods~\cite{BeyondAccuracy, DBLP:conf/acl/0013ZW0LS25} that focus on accuracy, our \textbf{Self-Critique} approach  targets confidence calibration. 
The model refines its confidence expression by analyzing the question, its reasoning steps, and final answer. But the results were unsatisfactory.
Thus, we propose \textbf{CritiCal}, a supervised fine-tuning (SFT) \textbf{Criti}que \textbf{Cal}ibration method that shifts from direct numerical optimization to critique-based learning~\cite{cft}.
During training, input consists of the question, model’s original response, and its confidence, while the output is a GPT-4o-generated critique of the confidence expression, based on the comparison of model’s reasoning process with a reference solution.
Additionally, we explore replacing SFT with direct preference optimization (DPO)~\cite{dpo} for the training of CritiCal, using GPT-4o critiques as chosen responses and the model’s Self-Critique as rejected ones due to its suboptimal performance.
Extensive experiments, both in-distribution and out-of-distribution, demonstrate that CritiCal significantly enhances confidence calibration for reasoning-intensive questions, \textit{surpassing even GPT-4o’s calibration capabilities}, as is shown in Figure~\ref{fig:Performance}. 
\textit{This suggests that a teacher model, with sufficient critique ability, can even enhance a student model’s confidence calibration beyond its own.}
CritiCal also exhibits robust calibration improvements in out-of-distribution settings, with models trained on critique-suited data even outperforming those trained in-distribution, highlighting its exceptional transferability.

\section{Related Works}

\subsection{Confidence Calibration}
Confidence calibration methods for LLMs are divided into white-box and black-box approaches. 
White-box methods use internal model data, such as attention mechanisms~\cite{attention, InternalState1}, hidden layers~\cite{InternalState}, or token probabilities~\cite{logit, ComparisonQA} for precise confidence estimates. 
Conversely, black-box ones rely on model outputs without accessing internal structure. 
Consistency-based methods assess confidence by sampling multiple outputs and measuring their similarity, assuming consistent responses indicate higher certainty~\cite{DBLP:journals/tmlr/LinT024, softlabel, consistency, consistency1}. 
Verbalization-based approaches train LLMs to explicitly express confidence through scores or epistemic markers~\cite{ConfTuner, marker, R-Tuning}. 
SaySelf~\cite{SaySelf} uses a teacher model to generate reflective rationales and confidence scores by analyzing inconsistencies across numerous sampled reasoning chains. 
However, it focuses on imitating the reference reasoning and confidence expressions rather than learning from critiques of its own confidence and is computationally inefficient due to reliance on diverse outputs.

\begin{figure*}[t]
\centering
\includegraphics[width=0.98\linewidth]{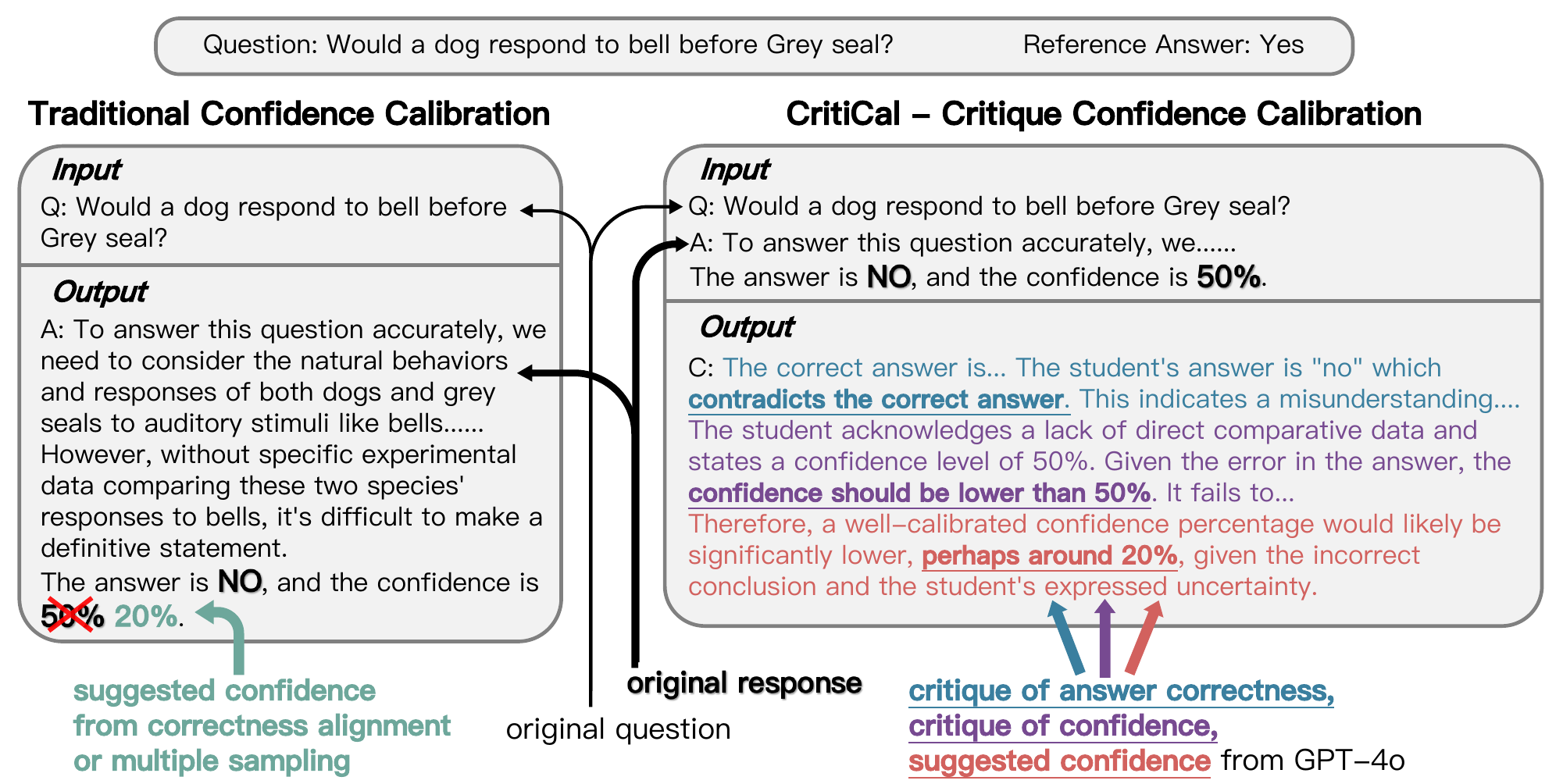} 
\vspace{-0.1in}
\caption{Comparisons between CritiCal and traditional confidence calibration methods.} 
\label{pipeline_fig}
\vspace{-0.2in}
\end{figure*}

\subsection{Critique Learning}
Self-correction has recently emerged as a promising approach to enhance LLMs' performance. 
Studies such as \citet{Self-Refine} and \citet{Welleck} utilize a model’s own feedback to refine outputs, though \citet{Huang}, \citet{curse} and \citet{Valmeekam} note limitations in its reliability for reasoning tasks. 
Alternatively, critique learning employs specialized models to provide feedback. 
\citet{critique_grpo} and \citet{Yang} develop outcome-based reward models, while \citet{Wang} and \citet{Lightman} focus on process-based reward models to improve reasoning by evaluating intermediate steps. 
\citet{BeyondBinary} uses critique to train LLMs to reason about their uncertainty but is still limited to numerical critiques. 
Further work by \citet{cft} explicitly leverages natural language critiques as a training objective to encourage deeper understanding and reasoning. 
However, it focuses on using critique to improve LLM accuracy, whereas our work explores natural language critiques to improve confidence calibration.

\section{Method}

To investigate whether critique can enhance confidence calibration, we propose two methods: \textbf{Self-Critique}, a prompting-based approach, and \textbf{CritiCal}, a supervised fine-tuning (SFT) framework.

\subsection{Self-Critique}

\label{sec: SC}

While prior studies on self-improvement~\cite{BeyondAccuracy, DBLP:conf/acl/0013ZW0LS25} focuses on refining reasoning processes and improving answer accuracy, \textbf{Self-Critique} targets confidence calibration, which aligns model’s verbalized confidence score with both the answer correctness and the uncertainty demonstrated in each reasoning steps. 
The model is prompted to reassess the question, its initial reasoning, and potential ambiguities or logical gaps, refining both the answer and confidence score to improve calibration.
The detailed prompt is provided in Appendix~\ref{sec:prompt}.

\subsection{CritiCal}

\label{pipeline}

To further enable LLMs to express well-calibrated confidence aligned with their reasoning, we propose \textbf{CritiCal}, a SFT method that guides LLMs to refine their confidence expressions using critiques of their initial confidence scores.

As illustrated in Figure~\ref{pipeline_fig}, CritiCal differs from traditional confidence calibration training methods~\cite{R-Tuning, SaySelf} in its input-output structure.
In conventional methods, input is the original question, and output is the original model answer paired with a suggested confidence expression, which is derived from either the alignment of answer correctness or the generation probability of such an answer during multiple times of response generation. 
In contrast, CritiCal is a sampling-free approach that encourages LLMs to learn from their confidence estimation errors through critique-based training. 
Specifically, the input consists of the question, the student model’s answer, and its associated confidence score, while the output is a critique from a teacher model (GPT-4o). 
This critique evaluates the calibration of the student’s confidence score, providing an explanation based on the clarity, strength, and correctness of the student’s reasoning compared to a reference solution.

In practice, we sample 2K questions from the training set and prompt the student model to generate answers along with confidence scores. 
These responses, paired with the questions and reference solutions from the benchmark, are provided to the teacher model to produce critiques assessing confidence calibration.
The student model is then fine-tuned using the collected critique data. 
In particular, to mitigate knowledge shift in large reasoning models (LRMs), we instruct the teacher model to structure their critiques with special "</think>" tokens, separating the explanation from the final judgment.  
This structured critique format facilitates more effective learning. 
The detailed prompt for critique generation is provided in Appendix~\ref{sec:prompt}.

\section{Experiments}

In this section, we answer the two questions: what to critique (\S\ref{sec: UaC}) and how to critique (\S\ref{sec: Self-Critique}, \S\ref{sec: CritiCal}).

\begin{figure*}[t]
\centering
\includegraphics[width=0.87\linewidth]{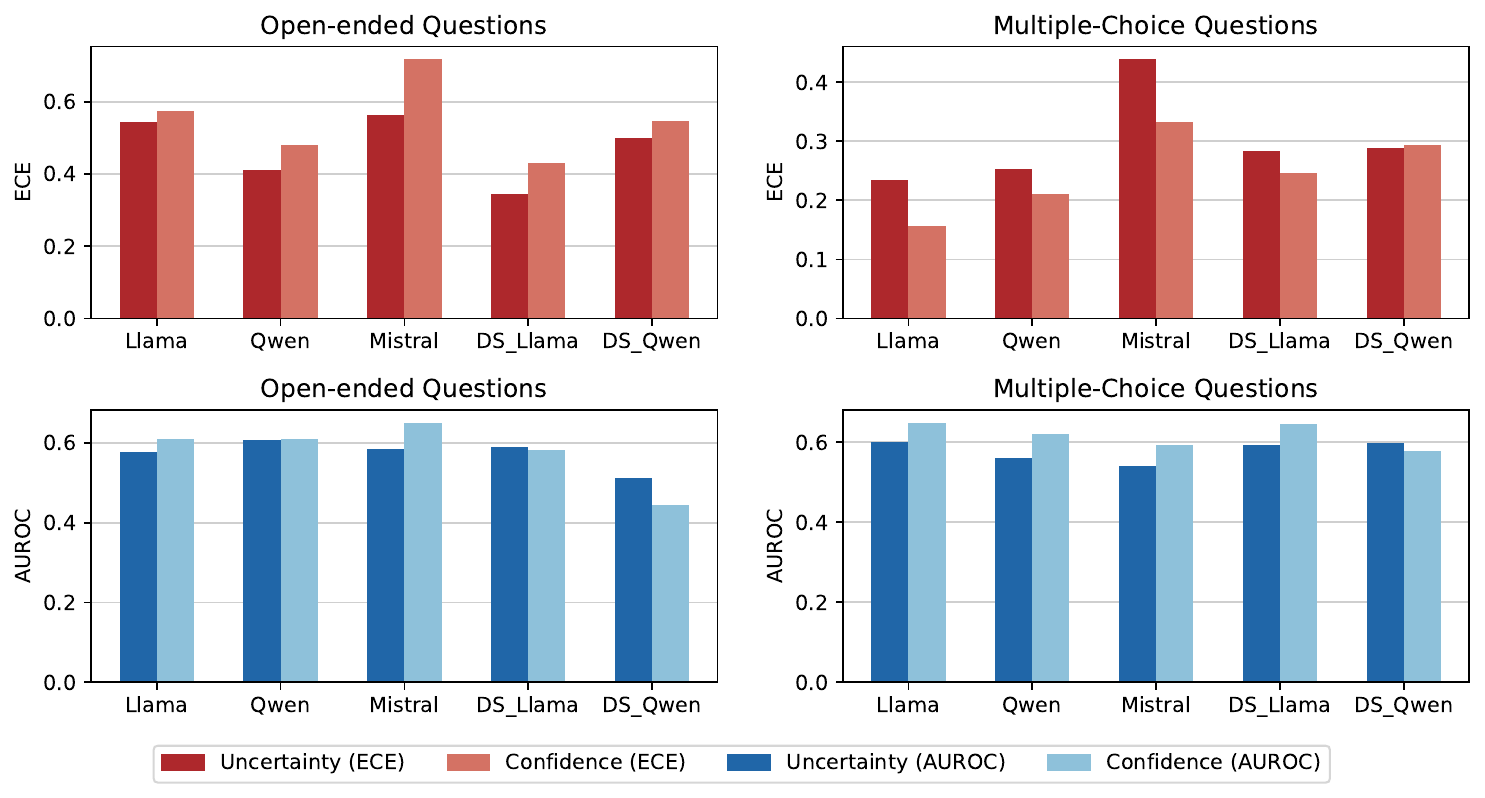}
\vspace{-0.1in}
\caption{Mean ECE and AUROC values for each model across the same category of benchmarks. The dark bars are the result under uncertainty prompt, and the light ones are of confidence. Further analysis under the setting of multi-turn Self-Critique can be found in Appendix~\ref{sec:Self-Critique_Results}.}
\label{task_unc_conf_comparison_pure}
\vspace{-0.2in}
\end{figure*}

\subsection{Experimental Setup}

\noindent\textbf{Datasets.}
All the experiments involved a total of 7 datasets: TriviaQA~\cite{TriviaQA} with open-ended, single-hop factuality questions; ComparisonQA~\cite{ComparisonQA} with multiple-choice, single-hop factuality questions; StrategyQA~\cite{StrategyQA} with yes/no, multi-hop factuality reasoning questions; HotpotQA~\cite{HotpotQA} with open-ended, multi-hop factuality reasoning questions; MATH~\cite{MATH} with open-ended, mathematical reasoning questions; MATH-500~\cite{MATH500} with harder ones selected from MATH test set; and MATH-Perturb~\cite{MATH-Perturb} with selected perturbed ones from MATH.

\noindent\textbf{Models.}
Our test involves LLMs: LLaMA~\cite{LLAMA3}, Qwen~\cite{qwen}, Mistral~\cite{Mistral}, LRMs: DeepSeek-Distill-Llama, DeepSeek-Distill-Qwen~\cite{DeepSeek}, and a proprietary API: GPT-4o~\cite{GPT4o}, for their diverse architectures.

\noindent\textbf{Metrics.}
We use accuracy (via exact match for open-ended questions) for response correctness measurement, expected calibration error (ECE) for confidence-accuracy alignment, and area under the receiver operating characteristic curve (AUROC) for confidence-based discrimination of correct and incorrect responses. 
For both accuracy and AUROC, the higher the better, but ECE is the opposite.

\subsection{Uncertainty vs. Confidence}

\label{sec: UaC}
Uncertainty, which pertains to the question as a whole, and confidence, which relates to the specific answer generated, are distinct concepts in LLMs, yet often mixed up in previous works~\cite{survey}. To address this, we investigate their differences and performance across various scenarios.

To ensure models distinguish between uncertainty and confidence, we provide clear definitions in the prompts, as detailed in Appendix~\ref{sec:prompt}. We evaluate five models across six benchmarks, grouped into open-ended and multiple-choice question types, with the MATH benchmark reserved exclusively for training.

Figure~\ref{task_unc_conf_comparison_pure} presents the mean ECE and AUROC for each model across benchmark categories.
The results reveal distinct performance patterns for uncertainty and confidence across question types. 
For open-ended questions, uncertainty consistently achieves lower ECE in both LLMs and LRMs. 
Conversely, for multiple-choice questions, confidence outperforms uncertainty in both ECE and AUROC.

This interesting discovery indicates that models exhibit better uncertainty calibration for open-ended questions, likely due to the expansive prediction space, where uncertainty captures the question’s inherent ambiguity. 
In contrast, confidence is better calibrated for multiple-choice questions, where the limited options allow models to leverage elimination strategies, enabling more precise confidence estimates for specific choices despite potential uncertainty about the question.

\begin{figure*}[t]
\centering
\includegraphics[width=0.98\linewidth]{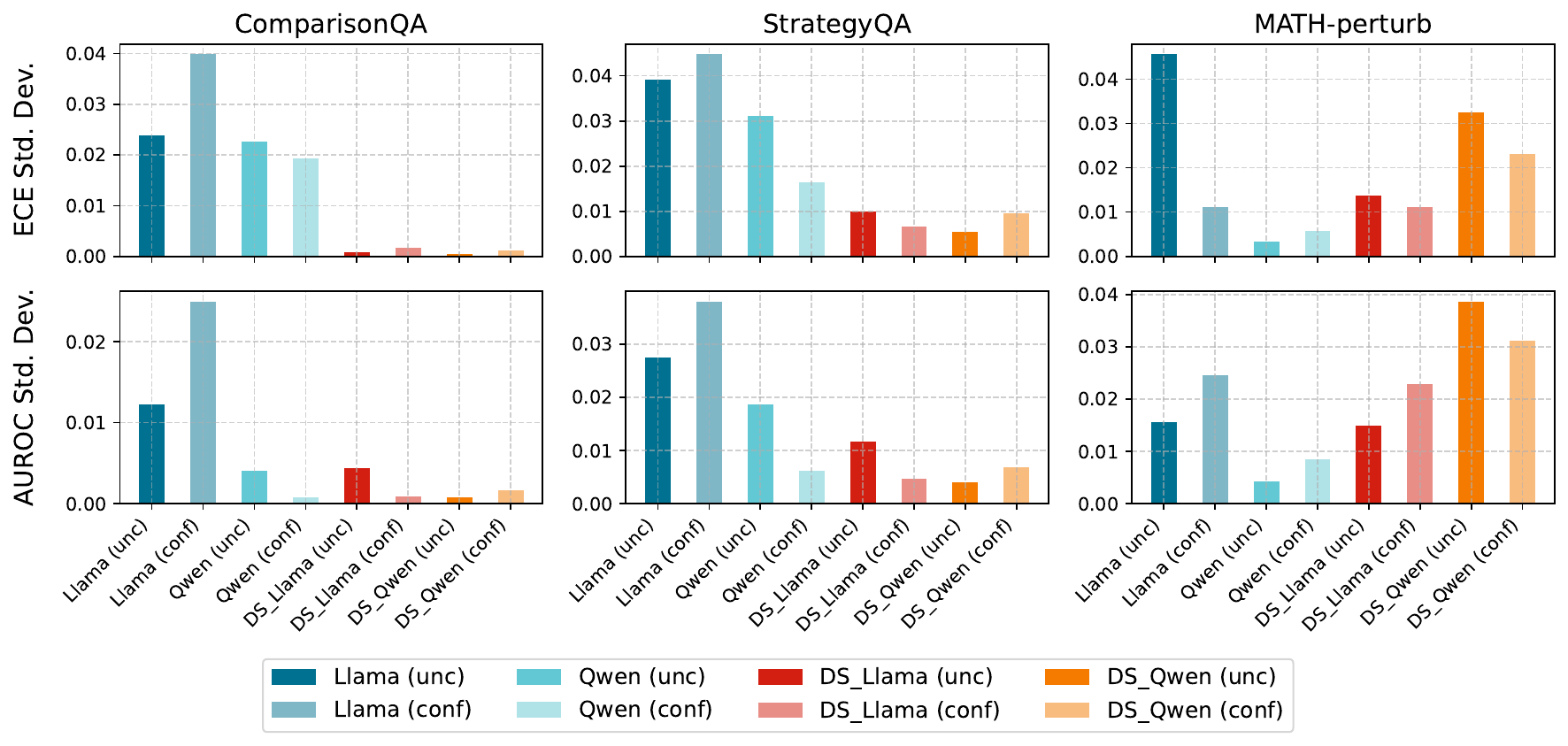}
\vspace{-0.1in}
\caption{Standard deviation of multi-turn Self-Critique for ECE and AUROC across three benchmarks. Each bar represents the standard deviation of a model's performance (uncertainty or confidence) across 6 iterations, where iteration 0 denotes the original response and iterations 1–5 indicate Self-Critique. Benchmarks are selected as representative of their task category due to question similarity under the same type. Full results are in Appendix~\ref{sec:Self-Critique_Results}.}
\label{all_datasets_combined_metrics}
\vspace{-0.1in}
\end{figure*}

\begin{figure}[t]
\centering
\includegraphics[width=0.98\linewidth]{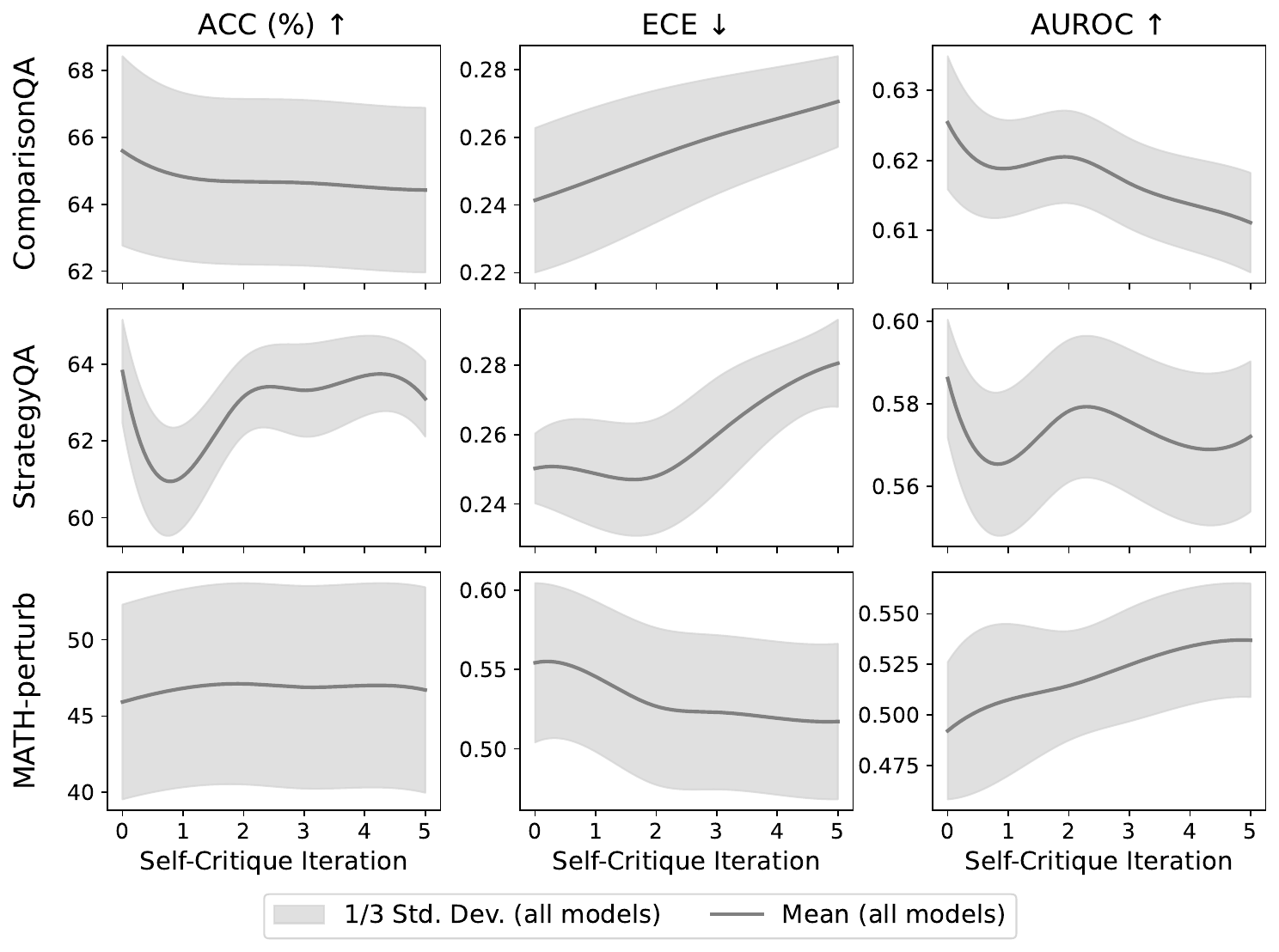}
\vspace{-0.1in}
\caption{Multi-turn Self-Critique results on ComparisonQA, StrategyQA, and MATH-perturb benchmarks. Each plot shows the smoothed mean performance (solid line) and the corresponding 1/3 standard deviation range (shaded area) for ACC, ECE, and AUROC. Iteration 0 represents the original response without Self-Critique.}
\label{all_datasets_combined_metrics_mean}
\vspace{-0.2in}
\end{figure} 

\subsection{Self-Critique Analysis}

\label{sec: Self-Critique}

This section investigates the impact of Self-Critique on confidence calibration and average confidence scores across multiple iterations.

\subsubsection{Multi-Turn Self-Critique}

\label{sec: MTSC}

To comprehensively evaluate Self-Critique performance, we conduct multi-turn Self-Critique experiments with 4 models, those have both reasoning and non-reasoning ones, across 6 benchmarks.
In each iteration, the model receives the results of all previous iterations as context. 
Detailed prompt is provided in Appendix~\ref{sec:prompt}.

Figure~\ref{all_datasets_combined_metrics} illustrates the standard deviation of multi-turn Self-Critique for each model, focusing on ECE and AUROC to evaluate confidence calibration stability. Figure~\ref{all_datasets_combined_metrics_mean} presents the average performance of all models across three benchmarks, highlighting the impact of Self-Critique on different tasks.
The specific variation curves of each model are shown in Figure~\ref{all_datasets_combined_metrics_full} in Appendix~\ref{sec:Self-Critique_Results}.

\noindent\textbf{Task Analysis.}
The six benchmarks are categorized into three tasks: one-hop factuality (ComparisonQA and TriviaQA), multi-hop factuality reasoning (StrategyQA and HotpotQA), and math reasoning (MATH500 and MATH-Perturb). 
As shown in Figure~\ref{all_datasets_combined_metrics_mean}, the semi-transparent light gray area represents the average performance of all models with a one-third standard deviation. 
For accuracy, models exhibit greater stability on one-hop factuality and math reasoning tasks compared to multi-hop factuality reasoning. 
However, unlike prior self-improvement studies~\cite{Self-Refine, Welleck}, Self-Critique shows no notable accuracy improvements, as it primarily targets confidence calibration rather than accuracy. 
For ECE and AUROC, Self-Critique exhibits relatively stable performance with slight improvements in math reasoning tasks.
In contrast, factuality-related benchmarks experience negative impacts, with increased average ECE and decreased average AUROC. 
These findings suggest that Self-Critique has limited effectiveness, significantly worsening calibration for factuality-related tasks while only marginally enhancing it for math reasoning. 
Thus, prompting-based Self-Critique alone is inadequate for robust confidence calibration.

\begin{figure*}[t]
\centering
\includegraphics[width=0.98\linewidth]{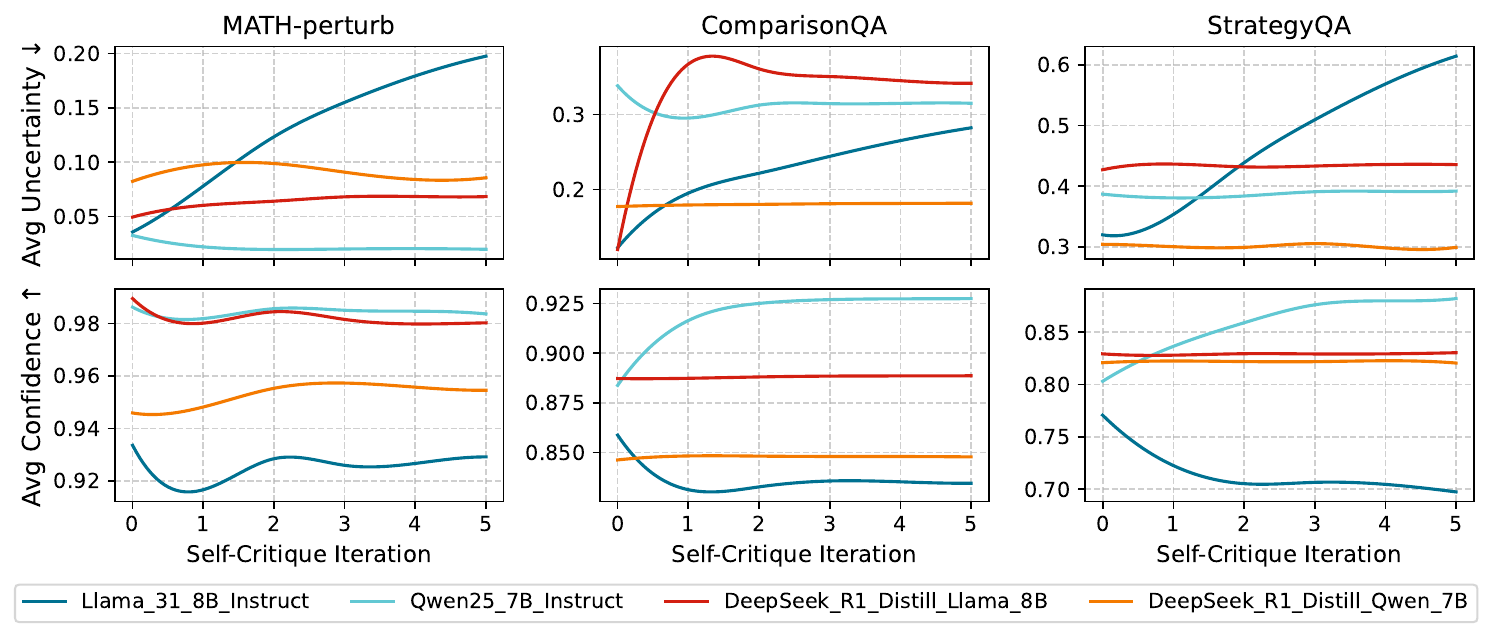}
\vspace{-0.1in}
\caption{Curves of average uncertainty and confidence scores during multi-turn Self-Critique across 3 benchmarks.} 
\label{task_avg_unc_conf_combined}
\vspace{-0.1in}
\end{figure*}

\noindent\textbf{Model Analysis.}
In Figure~\ref{all_datasets_combined_metrics} and Figure~\ref{all_datasets_combined_metrics_full}, LLMs are represented in cool colors, while LRMs are depicted in warm colors. 
For ECE and AUROC, LRMs demonstrate greater stability on factuality-related benchmarks, with significantly lower standard deviation than LLMs, whose calibration varies widely. 
This stability in LRMs probably arises from their extended reasoning processes, enabling deeper reflection on initial responses and preventing erratic confidence shifts.
Although LRMs show an increase in standard deviation in math-related questions, Figure~\ref{all_datasets_combined_metrics_full} reveals that this stems from their progressively refined confidence calibration. Overall, LRMs exhibit more consistent and reliable confidence calibration compared to LLMs.

\subsubsection{Average Confidence Change during Self-Critique}

Different from prior works~\cite{BeyondAccuracy} finding models become more confident despite incorrect answers during self-improvement, Self-Critique focuses on refining confidence calibration, leading to more complex outcomes as it prioritizes confidence expression over answer correctness.  

Results, shown in Figure~\ref{task_avg_unc_conf_combined}, vary significantly by model.
Llama consistently increases in uncertainty and decreases in confidence across all benchmarks, while Qwen shows the opposite trend, becoming more confident and less uncertain.
This suggests that multi-turn Self-Critique amplifies these model-specific tendencies. 
The two DeepSeek distilled models generally maintain more consistent uncertainty and confidence scores compared to non-reasoning models, except for an increase in uncertainty of the distilled Llama on ComparisonQA. 
This indicates that extended reasoning processes enhance the robustness of confidence expressions.

\subsection{CritiCal Analysis}

\label{sec: CritiCal}

We evaluate the performance of CritiCal in both in-distribution and out-of-distribution settings across multiple benchmarks.

We use the representative benchmark from each of the three tasks outlined in \S\ref{sec: MTSC}: one-hop factuality (ComparisonQA), multi-hop factuality reasoning (StrategyQA), and math reasoning (MATH-Perturb).
For fair comparison, we randomly sample 2K questions from the training set to construct training data each time, using the method described in \S\ref{pipeline}.
For MATH-Perturb, where questions are perturbations of a subset of MATH, we sample from the original MATH training set to build training data and test only on perturbed questions from the original test set to prevent data leakage. 
Training is conducted using LlamaFactory~\cite{LlamaFactory} with a batch size of 64 and other default hyper-parameters, taking approximately half an hour per each dataset on a 45G single GPU.

\subsubsection{In Distribution}

We first test the in-distribution performance of models fine-tuned with CritiCal, using Qwen and DeepSeek-Distill-Qwen as examples. 
Results are presented in Table~\ref{tab:cft}.

For fair comparison, we include several sampling-free baselines:
(1) \textbf{Vanilla}, a zero-shot prompt that directly asks model's verbalized confidence~\cite{CanLLMsExpress}. 
(2) \textbf{Self-Critique}, the non-training method described in \S\ref{sec: SC}. 
(3) \textbf{SFT\_Hard}, a SFT approach using a suggested confidence score based on model's original response (0\% for incorrect answers, 100\% for correct) for calibration~\cite{R-Tuning}, with uncertainty as the inverse.
(4) \textbf{SFT\_Soft}, a smoother SFT variant with confidence scores of 20\% and 80\%. 
(5) The performance of the teacher model, GPT-4o, is also included for reference.

\begin{table*}[t]
    \large
    \centering
    \resizebox{\linewidth}{!}{
	\begin{tabular}{@{}l|l|c||ccc|ccc|ccc@{}}
	\toprule
    \multirow{3}{*}{\textbf{Type}} &\multirow{3}{*}{\textbf{Method}}&\multirow{3}{*}{\textbf{Train}}&\multicolumn{3}{c|}{\textbf{ComparisonQA}} &\multicolumn{3}{c|}{\textbf{StrategyQA}}&\multicolumn{3}{c}{\textbf{MATH-Perturb}}
    \\
    &&&\textbf{ACC}&\textbf{ECE}&\textbf{AUROC}&\textbf{ACC}&\textbf{ECE}&\textbf{AUROC}&\textbf{ACC}&\textbf{ECE}&\textbf{AUROC}\\
    
    &&&\textbf{(↑)}&\textbf{(↓)}&\textbf{(↑)}&\textbf{(↑)}&\textbf{(↓)}&\textbf{(↑)}&\textbf{(↑)}&\textbf{(↓)}&\textbf{(↑)}\\
    
    \midrule
    
    \rowcolor[gray]{0.9} \multicolumn{12}{l}{\textbf{GPT-4o}} \\
    \midrule
    Uncertainty & Vanilla & N &  90.91 & 0.089 & 0.772  & 78.60 & 0.079 & 0.740 & 42.36 & 0.521 & 0.695\\
    Confidence & Vanilla & N &  91.97 & 0.036 & 0.787& 79.48 & 0.103 & 0.716  & 44.54 & 0.526 & 0.683 \\
    \midrule
    \rowcolor[gray]{0.9} \multicolumn{12}{l}{\textbf{Qwen-2.5-7B-Instruct (LLM)}} \\
    \midrule
    \multirow{5}{*}{Uncertainty} & Vanilla & N &  69.65 & \textbf{0.224} & 0.615& 64.63 & 0.283 & 0.507 &  39.57 & 0.587 & 0.525\\
    & Self\_Critique & N  & 68.24 & 0.268 & 0.605 & \textbf{67.25} & 0.308 & 0.464& 40.00 & 0.583 & 0.542\\
    & SFT\_Hard & Y &  69.49 & 0.229 & 0.616& 65.07 & 0.288 & 0.537 & 36.52& 0.605& 0.554\\
    & SFT\_Soft & Y & 69.68 & 0.228 & 0.615 & 64.19 & 0.245 & 0.564 & 38.70& 0.593& 0.558\\
    & \textbf{CritiCal} & Y & \textbf{69.76} & \textbf{0.224} & \textbf{0.619} & \textbf{67.25} & \textbf{0.221} & \textbf{0.597} & \textbf{40.87} & \textbf{0.558}& \textbf{0.586}\\
    \midrule
    
    \multirow{5}{*}{Confidence} & Vanilla  & N &  69.67 & 0.195 & 0.628& 65.07 & 0.226 & 0.612 & 37.83 & 0.609 & 0.571\\
    & Self\_Critique & N &  68.39 & 0.238 & \textbf{0.630} & 62.88 & 0.238 & 0.603& 37.39 & 0.610 & 0.578 \\
    & SFT\_Hard  & Y & 69.90 & 0.194 & 0.629 & 66.38 & 0.216 & 0.616 & \textbf{41.30} & 0.617 & 0.558\\
    & SFT\_Soft & Y & 69.90 & \textbf{0.193} & \textbf{0.630} & 66.38 &  0.193 & 0.629 & 38.70 & 0.611 & 0.562 \\
    & \textbf{CritiCal} & Y & \textbf{69.97} & 0.194 & \textbf{0.630}& \textbf{69.00} & \textbf{0.179} & \textbf{0.644} & 40.00 & \textbf{0.588} & \textbf{0.593}\\

    \midrule    
    \rowcolor[gray]{0.9} \multicolumn{12}{l}{\textbf{DeepSeek-R1-Distill-Qwen-7B (LRM)}} \\
    \midrule
    \multirow{5}{*}{Uncertainty} & Vanilla & N &  52.18 & 0.331 & 0.586& 58.52 & 0.247 & \textbf{0.609}  & 62.54 & 0.473 & 0.380\\
    & Self\_Critique  & N &  52.12 & 0.330 & \textbf{0.588}& 59.83 & 0.242 & 0.604 & 64.87 & 0.491 & 0.383 \\
    & SFT\_Hard & Y & \textbf{52.36} & \textbf{0.325} & 0.578&59.83 & 0.281 & 0.558 &65.65 & 0.446& 0.413\\
    & SFT\_Soft  & Y & 52.51 & \textbf{0.325} & 0.580& 62.45& 0.272 & 0.516 &66.09 &0.444 &0.437\\
    & \textbf{CritiCal} & Y & 52.30 & 0.326 & 0.579&  \textbf{65.07} & \textbf{0.223} & 0.572 & \textbf{67.83} & \textbf{0.405}& \textbf{0.457}\\
    \midrule
    
    \multirow{5}{*}{Confidence} & Vanilla  & N & 52.35 & \textbf{0.326} & 0.598 & 58.52 & 0.261 & 0.559 & 65.05 & 0.480 & 0.274\\
    & Self\_Critique & N &  52.31 & 0.327 & \textbf{0.602}& 57.64 & 0.278 & 0.541 & 65.05 & 0.516 & 0.271 \\
    & SFT\_Hard & Y & 52.62 & 0.332 & 0.577&  61.14 & 0.242 & 0.509 & 66.52 &0.487 & 0.270\\
    & SFT\_Soft & Y &\textbf{52.66}& 0.333 & 0.578& 61.14 &  0.235 & 0.597 & 65.65 & 0.467 & 0.301\\
    & \textbf{CritiCal}  & Y & 52.55 & 0.333 & 0.580 & \textbf{66.81} & \textbf{0.176} & \textbf{0.630} & \textbf{69.13} & \textbf{0.432} & \textbf{0.328}\\
    \bottomrule
    \end{tabular}
}
\vspace{-0.1in}
\caption{Performance of various LLMs and LRMs on ComparisonQA, StrategyQA, and MATH-Perturb. The "Train" column indicates whether the method needs additional training, providing a fair comparison. The best performances among all methods are \textbf{bold-faced}.} 
\label{tab:cft}
\vspace{-0.15in}
\end{table*}

Our key observations are as follows:
\textbf{(1) CritiCal excels in complex reasoning tasks.}
Although CritiCal shows limited impact on ComparisonQA, it significantly improves calibration and accuracy on StrategyQA and MATH-Perturb, showing a huge decrease in ECE and increase in AUROC compared to all baselines, including Self-Critique. 
This improvement stems from the long structured reasoning processes elicited by multi-hop and math reasoning tasks, which provide robust cues for critiquing confidence calibration.
\textbf{(2) CritiCal enables the student model to outperform even its teacher.}
Notably, on MATH-Perturb, GPT-4o further reduces the ECE of DeepSeek-Distill-Qwen, a model whose ECE is already lower than its. 
This demonstrates that a teacher model, with sufficient critique capabilities, can continuously enhance a student model’s confidence calibration, highlighting CritiCal’s potential.
\textbf{(3) Uncertainty and confidence distinctions persist in CritiCal.}
Models trained with CritiCal maintain the pattern observed in \S\ref{sec: UaC}: open-ended questions favor uncertainty, while multiple-choice questions favor confidence, as evidenced by superior ECE and AUROC performance, indicating what to critique.
\textbf{(4) Multi-hop factuality reasoning data is more suitable for critique than math reasoning.}
CritiCal yields greater calibration improvements on StrategyQA than on MATH-Perturb, suggesting that factuality reasoning questions, with their explicit multi-hop reasoning steps, are more critique-suited.

\subsubsection{Out of Distribution}

\begin{table*}[t]
    \large
    \centering
    \resizebox{\linewidth}{!}{
	\begin{tabular}{@{}l||ccc|ccc||ccc|ccc@{}}
	\toprule
    &\multicolumn{6}{c||}{\textbf{Uncertainty}}&\multicolumn{6}{c}{\textbf{Confidence}}
    \\
     \multirow{2}{*}{\textbf{Method}}&\multicolumn{3}{c|}{\textbf{In-distribution}}&\multicolumn{3}{c||}{\textbf{Out-of-distribution}}&\multicolumn{3}{c|}{\textbf{In-distribution}}&\multicolumn{3}{c}{\textbf{Out-of-distribution}}
    \\
&\textbf{ACC}&\textbf{ECE}&\textbf{AUROC}&\textbf{EM}&\textbf{ECE}&\textbf{AUROC}&\textbf{ACC}&\textbf{ECE}&\textbf{AUROC}&\textbf{EM}&\textbf{ECE}&\textbf{AUROC}\\
    &\textbf{(↑)}&\textbf{(↓)}&\textbf{(↑)}&\textbf{(↑)}&\textbf{(↓)}&\textbf{(↑)}&\textbf{(↑)}&\textbf{(↓)}&\textbf{(↑)}&\textbf{(↑)}&\textbf{(↓)}&\textbf{(↑)}\\
    
    \midrule
    \rowcolor[gray]{0.9} \multicolumn{13}{l}{\textbf{Qwen-2.5-7B-Instruct (LLM)}} \\
    \midrule
   SFT\_Hard &36.52& 0.605& 0.554 & 37.83 & 0.603 & 0.531& \textbf{41.30} & 0.617 & 0.558& 37.83 & 0.610 & 0.542\\
    SFT\_Soft  & 38.70& 0.593& 0.558 & 39.13 & 0.610 & 0.543 & 38.70 & 0.611 & 0.562& 36.09 & 0.625 & 0.578\\
    \textbf{CritiCal} & \textbf{40.87} & \textbf{0.558}& \textbf{0.586}& \textbf{39.57} & \textbf{0.574} & \textbf{0.595} & 40.00 & \textbf{0.588} & \textbf{0.593} & \textbf{42.17} & \textbf{0.571} & \textbf{0.593}\\
    
    
    \midrule
    \rowcolor[gray]{0.9} \multicolumn{13}{l}{\textbf{DeepSeek-R1-Distill-Qwen-7B (LRM)}} \\
    \midrule
    SFT\_Hard  & 65.65 & 0.446& 0.413 & 66.52 & 0.444 & 0.424 & 66.52 &0.487 & 0.270 & 64.78 & 0.493 & 0.266\\
    SFT\_Soft & 66.09 &0.444 &0.437 & 64.35 & 0.450 & 0.423 & 65.65 & 0.467 & 0.301 & 65.22 & 0.476 & 0.276\\
    \textbf{CritiCal} & \textbf{67.83} & \textbf{0.405}& \textbf{0.457}& \textbf{67.83} & \textbf{0.375} & \textbf{0.465} & \textbf{69.13} & \textbf{0.432} & \textbf{0.328} & \textbf{69.13} & \textbf{0.434} & \textbf{0.350}\\
    

    \bottomrule
    \end{tabular}
}
\vspace{-0.1in}
\caption{Comparisons of CritiCal's in-distribution and out-of-distribution performances. OOD Models are all trained on StrategyQA and tested on MATH-Perturb. 
The best performances among all methods are \textbf{bold-faced}.} 
\label{tab:ood}
\vspace{-0.08in}
\end{table*}

\begin{table}[h]
    \centering
    \resizebox{0.9\linewidth}{!}{
	\begin{tabular}{@{}c|c|ccc@{}}
	\toprule
    \textbf{Type}&\textbf{Method}&\textbf{ACC}&\textbf{ECE}&\textbf{AUROC}\\
    \midrule
    \rowcolor[gray]{0.9} \multicolumn{5}{l}{\textbf{StrategyQA (Multi-hop)}} \\
    \midrule
          \multirow{2}{*}{Uncertainty} & CFT & 67.25 & 0.221 & 0.597\\
           & CPO & 69.61 & 0.227 & 0.614\\
    \midrule
          \multirow{2}{*}{Confidence} & CFT &  69.00 & 0.179 & 0.644 \\
          & CPO & 66.81 & 0.181 & 0.634 \\
    \midrule
    
    \rowcolor[gray]{0.9} \multicolumn{5}{l}{\textbf{ComparisonQA (One-hop)}} \\
    \midrule
          \multirow{2}{*}{Uncertainty} & CFT & 69.76 & 0.224 & 0.619\\
          & CPO & 69.61 & 0.227 & 0.614\\
    \midrule
          \multirow{2}{*}{Confidence} & CFT & 69.97 & 0.194 & 0.630\\
          & CPO & 69.94 & 0.192 & 0.630 \\
		\bottomrule
	\end{tabular}
    }
\vspace{-0.02in}
\caption{Comparisons of using SFT and DPO as the training method respectively for CritiCal. } 
\label{tab:dpo}
\vspace{-0.1in}
\end{table}

To evaluate CritiCal’s generalization, we focus on its performance in out-of-distribution (OOD) settings~\cite{SurveySD, NewtonBench}.
Given CritiCal’s superior in-distribution performance on StrategyQA (as shown in Table~\ref{tab:cft}), we train models on StrategyQA and test them on MATH-Perturb for OOD analysis. Results are presented in Table~\ref{tab:ood}.

Both baseline fine-tuning methods (SFT\_Hard and SFT\_Soft) exhibit degraded performance on OOD data, indicating limited generalization. 
In contrast, CritiCal achieves improved calibration on OOD questions, with lower ECE and higher AUROC. 
This enhancement likely comes from StrategyQA’s critique-suited multi-hop reasoning data, which enables models to learn robust confidence calibration strategies based on their reasoning processes. 
These findings demonstrate CritiCal’s ability to foster reliable and generalizable confidence expressions across diverse tasks.

\subsubsection{Analysis of Training Method}

We also explore another popular optimization method, DPO~\cite{dpo}, for the training of CritiCal.
While the input structure remains identical, DPO differs in its output, consisting of a chosen response, the same as SFT’s output, and a rejected response. 
For the rejected response, which should have a similar structure to the chosen one, we use the model's Self-Critique output due to its suboptimal critique performance.

Since StrategyQA (multi-hop factuality reasoning) and MATH-Perturb (math reasoning) show similar performance trends in Table~\ref{tab:cft}, we test only StrategyQA for multi-hop reasoning due to limited computing resources.

For clarity, we denote SFT-based CritiCal as CFT and DPO-based CritiCal as CPO, with results shown in Table~\ref{tab:dpo}.
We can see that in both multi-hop and one-hop reasoning, the results of CFT and CPO differ very little compared to the huge improvement in Table~\ref{tab:cft}.
This suggests that CPO is also useful for reasoning-intensive tasks other than non-reasoning ones.
Given DPO’s higher computational cost, SFT remains a sufficient and efficient training method for CritiCal.

\section{Conclusions}

This study investigates critique-based learning to enhance verbalized confidence calibration in LLMs, addressing two key questions: 
(1) What to critique. 
Our findings reveal that confidence expressions are better suited for multiple-choice tasks, while uncertainty is more effective for open-ended tasks, providing clear guidance for calibration strategies. 
(2) How to critique. 
We introduced Self-Critique, which enables LLMs to refine their own confidence assessments, and CritiCal, a novel critique calibration method that leverages natural language critiques from a teacher model to optimize calibration.
Extensive experiments demonstrate that CritiCal significantly outperforms Self-Critique and other baselines, achieving superior calibration even beyond that of the teacher model, GPT-4o, in complex reasoning tasks. 
Moreover, CritiCal exhibits strong generalization, maintaining robust performance in both in-distribution and out-of-distribution settings, with notable transferability when trained on critique-suited multi-hop reasoning data.
And compared to DPO, SFT is sufficient and efficient for CritiCal training.
These findings underscore the potential of critique-based approaches to advance LLM reliability.

\section*{Limitations}
While CritiCal demonstrates significant improvements in confidence calibration for LLMs, several limitations still exist that cannot be covered in this single work.

The generalizability of CritiCal’s performance is  potentially constrained by the specific benchmarks used in our experiments. Although we select diverse tasks (one-hop factuality, multi-hop factuality reasoning, and math reasoning), these benchmarks may not fully represent the broad range of real-world scenarios where LLMs are deployed, such as creative writing or multi-modality tasks. Further evaluation on a wider array of datasets could strengthen claims about CritiCal’s robustness.

Additionally, computational constraints restrict our ability to evaluate all benchmarks in the comparison of training methods, SFT and DPO, where only ComparisonQA and StrategyQA are tested. Although these benchmarks are carefully chosen to represent one-hop and multi-hop factuality reasoning tasks, this limitation may obscure potential variations in CritiCal’s effectiveness across other task types. Future work could leverage greater computational resources to conduct a more comprehensive analysis, incorporating additional benchmarks and training configurations.

\section*{Ethics Statement}
This paper utilizes several publicly available datasets, including ComparisonQA, TriviaQA, StrategyQA, HotpotQA, MATH, MATH-Perturb, and MATH-500, which are accessible to the research community under CC, Apache 2.0, MIT, Apache 2.0, MIT, MIT, and MIT licenses, respectively.
The data is anonymized, ensuring our work does not raise privacy concerns regarding specific entities.

Our experiments involve the use of LLaMA, Qwen, Mistral, DeepSeek-Distill-Llama, DeepSeek-Distill-Qwen, and GPT-4o, so the same risks from LLMs research are also applicable to this work.

While CritiCal seeks to increase trust in AI by training on natural language critiques, there is a risk of users overly relying on its confidence estimates. These estimates may occasionally be inaccurate. Therefore, users are advised to treat these confidence expressions as a reference only.

\section*{Acknowledgments}

The authors of this paper were supported by the ITSP Platform Research Project (ITS/189/23FP) from ITC of Hong Kong, SAR, China, and the AoE (AoE/E-601/24-N), the RIF (R6021-20) and the GRF (16205322) from RGC of Hong Kong,SAR, China.


\bibliography{custom}

\newpage
\appendix

\begin{center}
    {\Large\textbf{Appendices}}
\end{center}

\section{Prompt}
\label{sec:prompt}

We design our prompt according to previous works~\cite{CanLLMsExpress, gprooft, Prospect}. Figure~\ref{fig:pure_unc_prompt}, \ref{fig:pure_conf_prompt}, \ref{fig:self_critique_conf_prompt}, \ref{fig:gen_critique_prompt} illustrate the prompt we use for vanilla uncertainty inquiry, vanilla confidence inquiry, Self-Critique with confidence, and critique generation, respectively, with StrategyQA as an example.

\begin{figure}[h!]
\begin{tcolorbox}[
    colback=gray!3!white,
    colframe=black!30!white, 
    title=Vanilla prompt using uncertainty,
    fonttitle=\bfseries,
    boxrule=0.5pt,
    arc=4pt,
    boxsep=5pt,
    left=6pt,
    right=6pt,
    top=6pt,
    bottom=6pt,
    coltitle=blue!50!black
]
Answer the following yes/no question and provide your uncertainty score. Your response should end with 'The answer is [your\_answer], and the uncertainty is [uncertainty\_percentage]\%' where [your\_answer] is yes or no, and the uncertainty percentage is a number between 0 and 100, indicating how uncertain you are about the question. If you are not sure, you should give a higher uncertainty percentage.

Question: \textit{[Question]}
\end{tcolorbox}
\caption{The vanilla prompt using uncertainty on StrategyQA. Placeholders \textit{[Question]} will be replaced with the real one.}
\label{fig:pure_unc_prompt}
\end{figure}

\begin{figure}[h!]
\begin{tcolorbox}[
    colback=gray!3!white,
    colframe=black!30!white, 
    title=Vanilla prompt using confidence,
    fonttitle=\bfseries,
    boxrule=0.5pt,
    arc=4pt,
    boxsep=5pt,
    left=6pt,
    right=6pt,
    top=6pt,
    bottom=6pt,
    coltitle=blue!50!black
]
Answer the following yes/no question and provide your confidence score. Your response should end with 'The answer is [your\_answer], and the confidence is [confidence\_percentage]\%' where [your\_answer] is yes or no, and the confidence percentage is a number between 0 and 100, indicating how sure you are about your answer. If you are not sure, you should give a lower confidence percentage.

Question: \textit{[Question]}
\end{tcolorbox}
\caption{The vanilla prompt using confidence on StrategyQA. Placeholders \textit{[Question]} will be replaced with the real one.}
\label{fig:pure_conf_prompt}
\end{figure}

\begin{figure}[h!]
\begin{tcolorbox}[
    colback=gray!3!white,
    colframe=black!30!white, 
    title=Multi-turn Self-Critique prompt using confidence on StrategyQA,
    fonttitle=\bfseries,
    boxrule=0.5pt,
    arc=4pt,
    boxsep=5pt,
    left=6pt,
    right=6pt,
    top=6pt,
    bottom=6pt,
    coltitle=blue!50!black
]
You previously answered the following yes/no question, and your responses have gone through one or more rounds of refinement. Below is the question, your initial response, and all subsequent refined responses. Now, reassess the question and your previous reasoning, including the initial and all refined responses. Consider any potential ambiguities, logical steps, or overlooked aspects that could improve the accuracy of your response and the calibration of your confidence score. Answer the question and provide a new confidence score.

Question: \textit{[Question]}

Initial response: \textit{[Initial\_Responses]}

Refined responses: \textit{[Refined\_Responses]}

Your response should end with 'The refined answer is [your\_answer], and the confidence is [confidence\_percentage]\%' where [your\_answer] is yes or no, and the confidence percentage is a number between 0 and 100, indicating how sure you are about your refined answer. If you are not sure about your refined answer, you should give a lower confidence percentage.
\end{tcolorbox}
\caption{The prompt for multi-turn Self-Critique using confidence on StrategyQA. Placeholders \textit{[Question]}, \textit{[Initial\_Responses]}, and \textit{[Refined\_Responses]} will be replaced with the real ones.}
\label{fig:self_critique_conf_prompt}
\end{figure}

\begin{figure}[h!]
\begin{tcolorbox}[
    colback=gray!3!white,
    colframe=black!30!white, 
    title=Critique generation prompt on StrategyQA,
    fonttitle=\bfseries,
    boxrule=0.5pt,
    arc=4pt,
    boxsep=5pt,
    left=6pt,
    right=6pt,
    top=6pt,
    bottom=6pt,
    coltitle=blue!50!black
]
Confidence indicates how how sure the student is about his answer. If he is not sure, he should give a lower confidence percentage. You are a teacher expert in confidence calibration. A student previously answered a question and provided his confidence score. Please evaluate the calibration of his confidence score for the question based on his response. If his response is incorrect, the confidence percentage should be low. 

Question: \textit{[Question]}

Correct Answer: \textit{[Correct\_Answer]}

Facts: \textit{[Facts]}

Student's Response: \textit{[Student's\_Response]}

Using the facts and the correct answer as a reference, assess whether the confidence percentage in the student's response is well-calibrated, considering the clarity and strength of the reasoning provided and your own knowledge of the question. Is the confidence percentage appropriate, too high, or too low? Provide a brief explanation of your evaluation, focusing on how well his confidence aligns with the strength of his reasoning and the context of the question.
\end{tcolorbox}
\caption{The prompt we use to generate confidence calibration critique on StrategyQA. Placeholders \textit{[Question]}, \textit{[Correct\_Answer]}, \textit{[Facts]}, and \textit{[Student's\_Response]} will be replaced with the real ones.}
\label{fig:gen_critique_prompt}
\end{figure}






\section{Detailed Self-Critique Results}
\label{sec:Self-Critique_Results}

\begin{figure*}[h]
\centering
\includegraphics[scale=0.45]{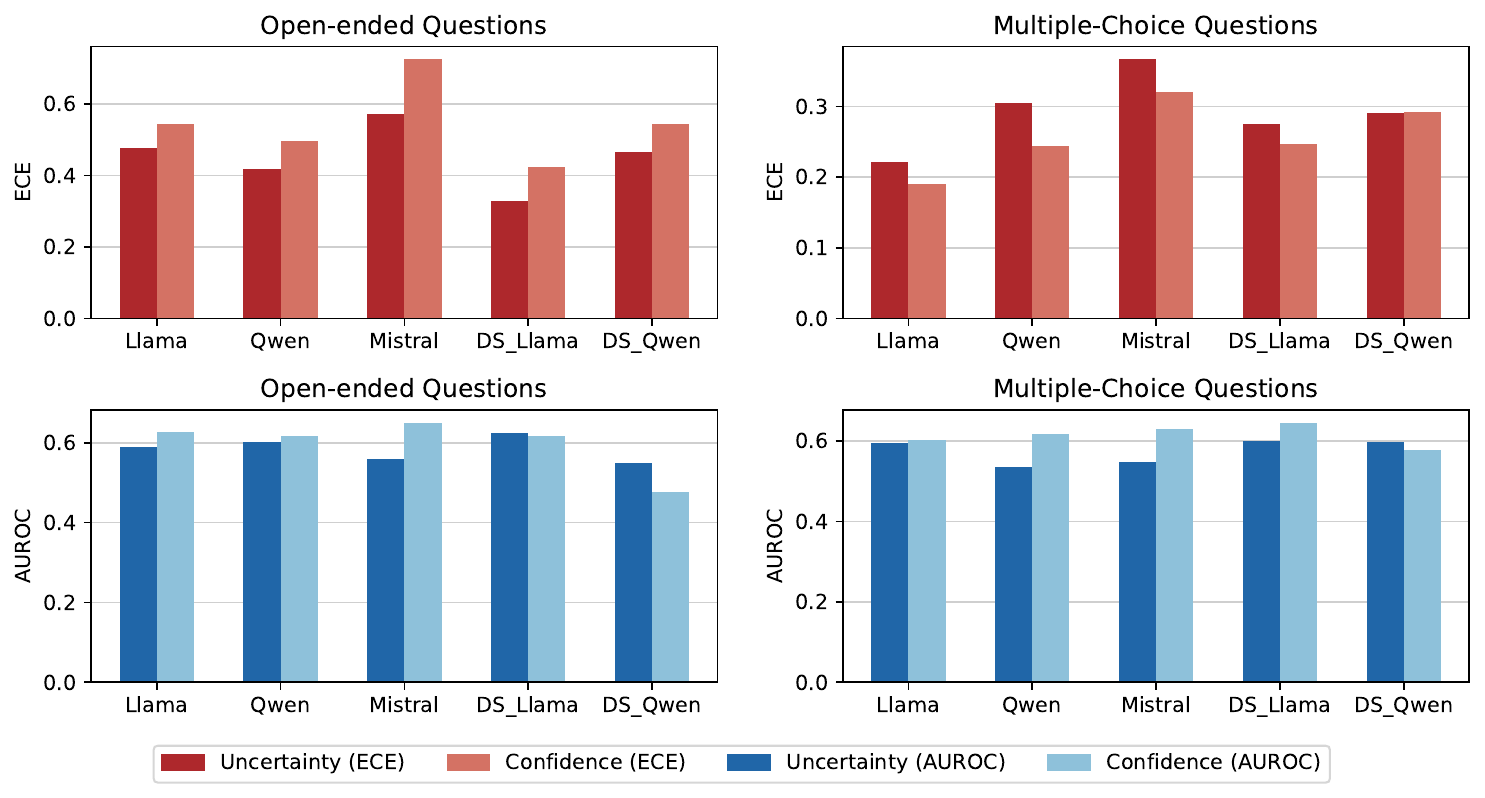}
\vspace{-0.15in}
\caption{Mean ECE and AUROC values for each model across the same category of benchmarks, which are taken the average of across the 5 turns of Self-Critique. The dark bars are the result under uncertainty prompt, and the light ones are of confidence.} 
\label{task_unc_conf_comparison}
\vspace{-0.2in}
\end{figure*}

Figure~\ref{task_unc_conf_comparison} shows the difference between uncertainty and confidence after Self-Critique. The distinctions between these two concepts still remain evident after applying Self-Critique.

\begin{figure*}[t]
\centering
\includegraphics[width=0.75\linewidth]{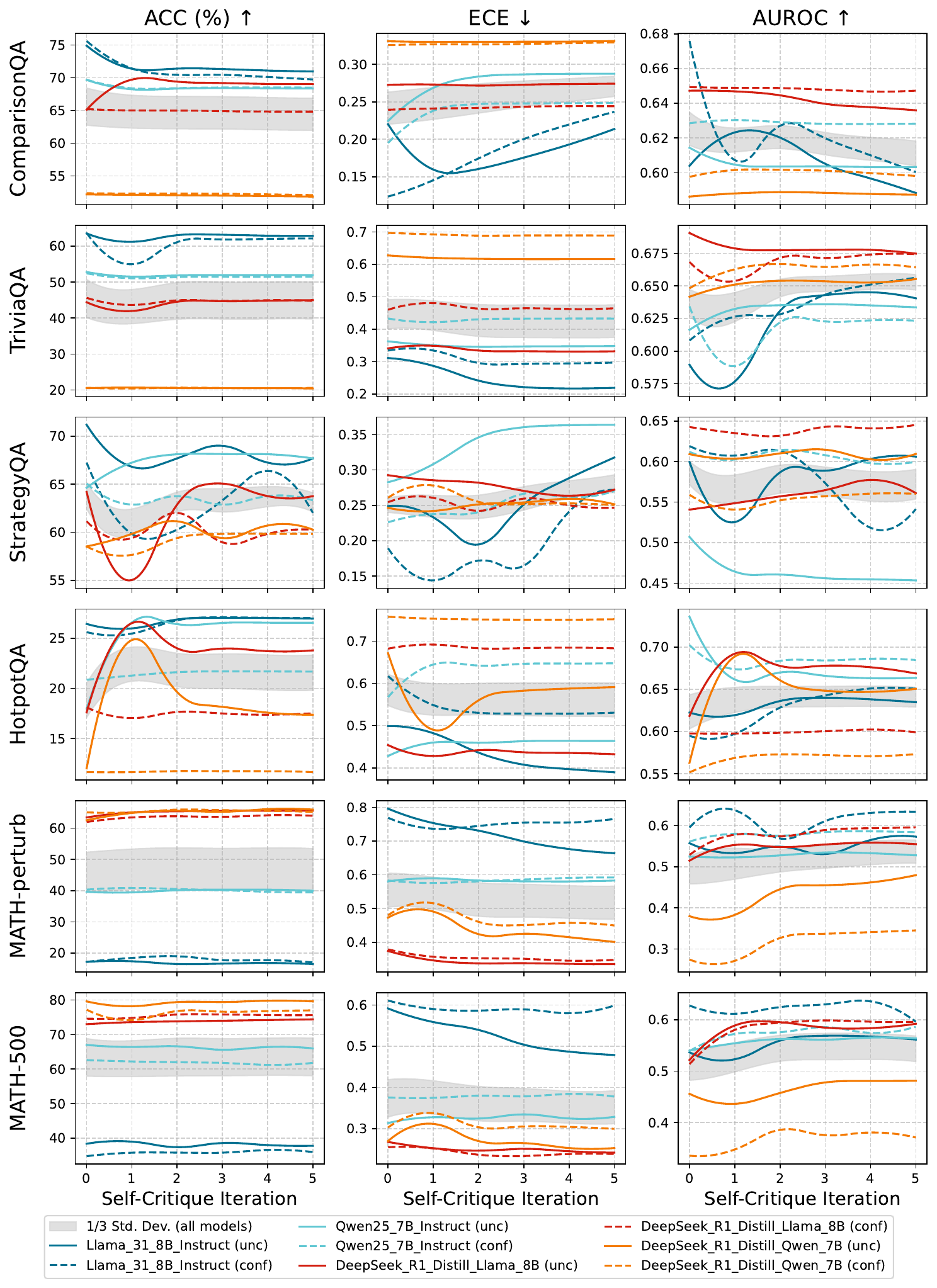}
\vspace{-0.15in}
\caption{Multi-turn Self-Critique results on all the six benchmarks. The 0 iteration means the original response without Self-Critique. The semi-transparent light gray area represents the average performance of all models with a one-third standard deviation.} 
\label{all_datasets_combined_metrics_full}
\vspace{-0.2in}
\end{figure*}

Figure~\ref{all_datasets_combined_metrics_full} displays the performance trajectories of each model across all six benchmarks. Self-Critique demonstrates relatively stronger improvements on mathematical reasoning tasks but falls short on factuality-related tasks, highlighting its limitations and lack of robustness.

\end{document}